\documentclass[letterpaper, 10 pt, conference]{IEEEtran}  

\IEEEoverridecommandlockouts                              

\usepackage{graphicx} 
\usepackage{caption}
\usepackage{subcaption}
\usepackage{latexsym}
\usepackage{multirow}

\usepackage{hyperref}
\hypersetup{colorlinks}

\usepackage{amsmath}
\newcommand{\norm}[1]{\left\lVert#1\right\rVert}

\usepackage{booktabs}
\usepackage{placeins}

\usepackage{url}
\title{\LARGE \bf
Forward-Looking Sonar Patch Matching: Modern CNNs, Ensembling, and Uncertainty
}

\author{Arka Mallick$^{1}$ and Paul Pl\"oger$^{1}$ and Matias Valdenegro-Toro$^{2}$
\thanks{$^{1}$Department of Computer Science, Hochschule Bonn-Rhein-Sieg, 53757 Sankt Augustin, Germany
        {\tt\small arkamallick30@gmail.com}}%
\thanks{$^{2}$German Research Center for Artificial Intelligence, 28359 Bremen, Germany
        {\tt\small matias.valdenegro@dfki.de}}%
}

\begin{document}

\maketitle
\thispagestyle{empty}
\pagestyle{empty}

\begin{abstract}
Application of underwater robots are on the rise, most of them are dependent on sonar for underwater vision, but the lack of strong perception capabilities limits them in this task.
An important issue in sonar perception is matching image patches, which can enable other techniques like localization, change detection, and mapping. There is a rich literature for this problem in color images, but for acoustic images, it is lacking, due to the physics that produce these images.
In this paper we improve on our previous results for this problem (Valdenegro-Toro et al, 2017), instead of modeling features manually, a Convolutional Neural Network (CNN) learns a similarity function and predicts if two input sonar images are similar or not. With the objective of improving the sonar image matching problem further, three state of the art CNN architectures are evaluated on the Marine Debris dataset, namely DenseNet, and VGG, with a siamese or two-channel architecture, and contrastive loss. To ensure a fair evaluation of each network, thorough hyper-parameter optimization is executed. 
We find that the best performing models are DenseNet Two-Channel network with 0.955 AUC, VGG-Siamese with contrastive loss at 0.949 AUC and DenseNet Siamese with 0.921 AUC. By ensembling the top performing DenseNet two-channel and DenseNet-Siamese models overall highest prediction accuracy obtained is 0.978 AUC, showing a large improvement over the 0.91 AUC in the state of the art.

\end{abstract}

\section{Introduction}

More than two-thirds of our planet's surface is covered by oceans and other water bodies. For a human, it is often impossible to explore it extensively. 
The need for venturing into potentially dangerous underwater scenarios appear regularly, for example, finding new energy sources, monitoring tsunamis, global warming, wreckage search, or maybe just to learn about deep sea ecosystems. This motivates design and deployment of robots in underwater scenarios, and much research goes in this direction.
Some exploration or monitoring tasks require the robot to "see" underwater, to make intelligent decisions.
But the underwater environment is very difficult for optical cameras, as light is attenuated and absorbed by the water particles. And a lot of real-life monitoring and mapping tasks take place in a cluttered and turbid underwater scenario. The limited visibility range of an optical sensor is a big challenge. Hence, sonar is a more practical choice for underwater sensing, as acoustic waves can travel long distances with comparatively little attenuation.

An underwater robot, equipped with sonar image sensors, regularly needs to perform basic tasks such as object detection and recognition, navigation, manipulation etc. In underwater scenarios, sonar patch matching functionality is very useful in several applications such as data association in simultaneous localization and mapping (SLAM), object tracking, sonar image mosaicing \cite{hurtos2012fourier} etc. Patch matching, in general, is heavily used in computer vision and image processing applications for low-level tasks like image stitching \cite{brown2007automatic}, 
deriving structure from motion \cite{molton2004locally}, also in high-level tasks such as object instance recognition \cite{lowe1999object}, object classification \cite{yao2012codebook}, multi-view reconstruction \cite{seitz2006comparison}, image-retrieval etc. 

Typical challenges in patch matching tasks are different viewing points, variations in scene insonification, occlusion, and different sensor settings. For sonar patch matching the common challenges with acoustic vision adds to the overall complexity. For example, low signal-to-noise ratio, lower resolution, unwanted reflections, less visibility etc. 
Because of these challenges, the underlying object features might not be so prominent as in a normal optical image. It has also been found that it is very challenging to manually design features for sonar images, and popular hand designed features such as SIFT \cite{lowe2004distinctive} are not always very effective in sonar images \cite{stateoftheart}. 
For these reasons, patch matching for sonar images remains a topic of research interest.

\begin{figure}[t]
    \centering
    \includegraphics[width=0.2\textwidth]{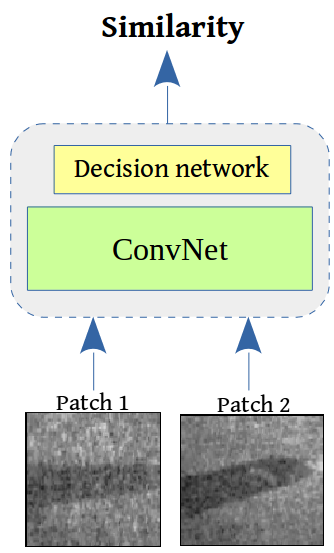}
    \caption{Use of convolutional network for learning general similarity function for image patches. The patches in the image are samples taken from the data used in this work. 
        Inspired from Zagoruyko et al. \cite{zagoruyko2015learning}}
    \label{similarity_function_wraped}
\end{figure} 

\section{State of the Art}
Sonar image patch matching is more difficult than normal optical matching problem. This is because sonar images have additional challenges such as non-uniform insonification, low signal-to-noise ratio, poor contrast \cite{emberton2018underwater}, low resolution, low feature repeatability \cite{hurtos2013automatic} etc. But sonar image matching has important applications like in sonar registration, mosaicing \cite{kim2005mosaicing}, \cite{hurtos2012fourier} and mapping of seabed surface \cite{negahdaripour2011dynamic} etc. 
While Kim et al. \cite{kim2005mosaicing} used Harris corner detection and matched key-points to register sonar images, Hurtos et al. \cite{hurtos2012fourier} incorporated Fourier-based features for registration of FLS images. Negahdaripour
et al. \cite{negahdaripour2011dynamic} estimated mathematical models from the dynamics of object movements and it's shadows. Vandrish et al. \cite{vandrish2011side} used SIFT \cite{lowe2004distinctive} for sidescan sonar image registration.
Even though these approaches achieve considerable success in respective goals, were found to be most effective when the rotation/translation between the frames of sonar images are comparatively smaller. Block-matching was performed on segmented sonar images by Pham et al. \cite{pham2013guided}, using Self-Organizing Map for the registration and mosaicing task.

Recently CNNs have been applied for this problem, Zbontar et al\cite{zbontar2016stereo} for stereo matching in color images, and Valdenegro-Toro et al \cite{stateoftheart} for sonar images, which is based on Zagoryuko et al \cite{zagoruyko2015learning}, and is the state of the art for sonar image patch matching at 0.91 AUC on the Marine Debris dataset. CNNs are increasingly being used for sonar image processing \cite{valdenegro2016objectness}. The main reason behind such a rise of CNN usage is that it can learn sonar-specific information from the data directly. No complex manual feature design or rigorous data pre-processing steps are needed, which makes the task less complex and good prediction accuracy can be achieved.

\section{Matching as Binary Classification}

We formulate the matching problem as learning a classifier. A classification model is given two images, and it decides if the images match or not. This decision can be modeled as a score in $y \in [0, 1]$, or a binary output decision $y = 0, 1$.

For this formulation, we use AUC, the area under the ROC curve (Receiver Operating Characteristic) as the primary metric to assess performance, as we are interested in how separable are the score distributions between matches and non-matches.

\begin{table}[t]
    \centering
    \caption{Best hyper-parameter values for DenseNet Two-Channel (DTC).}
    \resizebox{0.45\textwidth}{!}
    {
        \begin{tabular}{llll} 
            \toprule
            \multicolumn{1}{l}{\textbf{Name}} & \multicolumn{1}{l}{\textbf{Value}} & \multicolumn{1}{l}{\textbf{Name}} & \multicolumn{1}{l}{\textbf{Value}} \\ [0.5ex] 
            \midrule
            Layers & 2-2-2 & Pooling & avg\\
            Growth rate (gr) & 12 & Number of filter & 32\\ 
            DenseNet dropout & 0.2 & Compression & 0.5 \\ 
            Bottleneck & False & Batch size & 128 \\
            Optimizer & Adadelta & Learning rate & 0.03 \\
            \toprule
        \end{tabular}
    }
    \label{table:final_run_search_space_dtc}
\end{table}

\section{Matching Architectures}

In this section we describe the neural network architectures we selected as trunk for the meta-architectures like two-channel and siamese networks, which are used for matching.

\subsection{Hyper-Parameter Tuning}

For each architecture, we tuned their hyper-parameters using a validation set, in order to maximize accuracy. Each range of hyper-parameters was set individually for each architecture, considering width, filter values at each layer, drop probabilities, dense layer widths, etc. Overall, we performed 10 runs of different hyper-parameter combinations for each architecture. Details of the hyper-parameter tuning are available at \cite{mallick2019thesis}.

\subsection{DenseNet Two-Channel Network}
In DenseNet \cite{densenet} each layer connects to every layer in a feed-forward fashion. With the basic idea to enhance the feature propagation, each layer of DenseNet blocks takes the feature-maps of the previous stages as input.  


\begin{table}[t]
    \centering
    \caption{Best hyperparameter values for DenseNet Siamese (DS).}
    \resizebox{0.4\textwidth}{!}
    {\begin{tabular}{llll} 
            \toprule
            \multicolumn{1}{l}{\textbf{Name}} & \multicolumn{1}{l}{\textbf{Value}} & \multicolumn{1}{l}{\textbf{Name}} & \multicolumn{1}{l}{\textbf{Value}} \\ [0.5ex] 
            \midrule
            Number of filter & 16 & Layers & 2-2 \\
            Growth rate & 30 & DenseNet dropout & 0.4\\
            Compression & 0.3 & Bottleneck & False\\
            FC output & 512 & FC dropout & 0.7 \\
            Pooling & flatten & Batch size & 64 \\
            Optimizer & Adadelta & Learning rate & 0.07\\
            \toprule
    \end{tabular}}
    \label{table:final_run_search_space_dns}
\end{table}

In DenseNet two channel the the sonar patches are supplied as inputs in two channels format, the network by itself divides each patch into one channel and learn the features from the patches and then finally compare them using the Sigmoid activation function at the end with FC layer of single output. 

Hyper-parameters for this architecture are shown in Table \ref{table:final_run_search_space_dtc}.


\subsection{DenseNet Siamese Network}
In this architecture the branches of the Siamese network are DenseNet. Following the classic Siamese model each branch of the Siamese network shares weights between them and gets trained simultaneously on two input patches and then learns the features from the inputs.
Through the shared neurons the Siamese network is able to learn the similarity function and be able to discriminate between the two input patches. The role of the DenseNet branches are feature extraction, the decision making or prediction 
part is taken care of by the Siamese network.

\begin{figure}[t]
    \centering
    \includegraphics[width=0.35\textwidth]{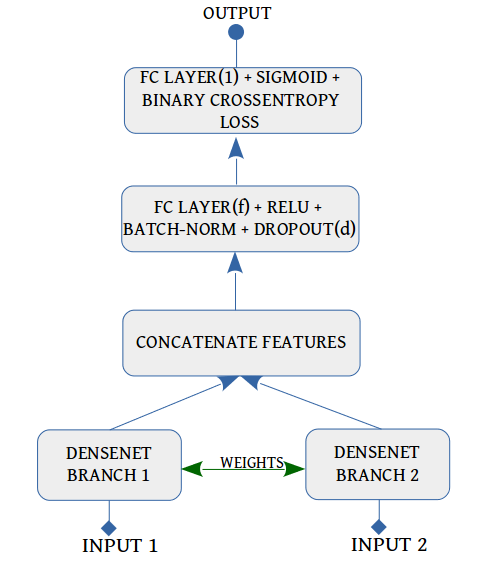}
    \caption{DenseNet Siamese architecture.}
    \label{fig:dn_siamese}
\end{figure}

In Figure \ref{fig:dn_siamese} the basic architecture is displayed for the DenseNet-Siamese network. The two DenseNet branches are designed to share weights between them. The extracted features are concatenated and connected through a FC layer, followed by ReLU activation and where applicable Batch Normalization and Dropout layers. The output is then connected to another FC layer with single output, for binary prediction score of matching (1) or non-matching (0). Sigmoid activation function and binary cross entropy loss function is used for this final FC layer. As mentioned in Figure \ref{fig:dn_siamese} the size of the output of the FC layer and value of dropout probability etc. hyper-parameters are shown in Table \ref{table:final_run_search_space_dns}.

\subsection{Contrastive Loss}
Using Contrastive loss \cite{hadsell2006dimensionality} higher dimensional input data (e.g. a pair of images) can be mapped in a much lower dimensional output manifold, where similar pairs are placed closer to each other and the dissimilar pairs have larger distances between them depending on their dissimilarity. Using this loss function the distance between two input patches projected in the output manifold can be predicted and if the distance is closer to 0 then the input pairs are matching, otherwise its dissimilar (above threshold). The formulas for this loss are shown in Equations \ref{eq:distance} and \ref{eq:contrastive}.
\begin{equation}
    D_{W}(\vec{X}_{1}, \vec{X}_{2}) = \norm{ G_{W}(\vec{X}_{1}) - G_{W}(\vec{X}_{2})}
    \label{eq:distance}
\end{equation}
\begin{equation}
    L = (1-Y) \frac{1}{2} (D_{W})^{2} + \frac{Y}{2} \{\max(0, m-D_{W})\}^{2}
    \label{eq:contrastive}
\end{equation}
Here L is the loss term, the formula presented here is the most generalized form of the loss function, suitable for batch training. 
$ \vec{X_1}$, $ \vec{X_2}$ represents a pair of input image vectors. Y are the labels, 0 for similar pair and 1 for dissimilar pair. $D_w$ is the parameterized distance function to be learned by the neural network.
$m > 0$ is the margin that defines a radius around $G_w$. The dissimilar pairs only contribute to the loss function if their distance is within the radius. We use $m = 1$ for our experiments.
One of the ideas for evaluating this loss function is to use it with a Siamese network, as the loss function takes a pair of images as input, indicating their similarity, matching pairs having closer distances in the learned embedding than non-matching ones, and the distance between pairs can be used as a score with a threshold.

\subsection{VGG Siamese Network}

\begin{figure}[t]
    \centering
    \includegraphics[width=0.32\textwidth]{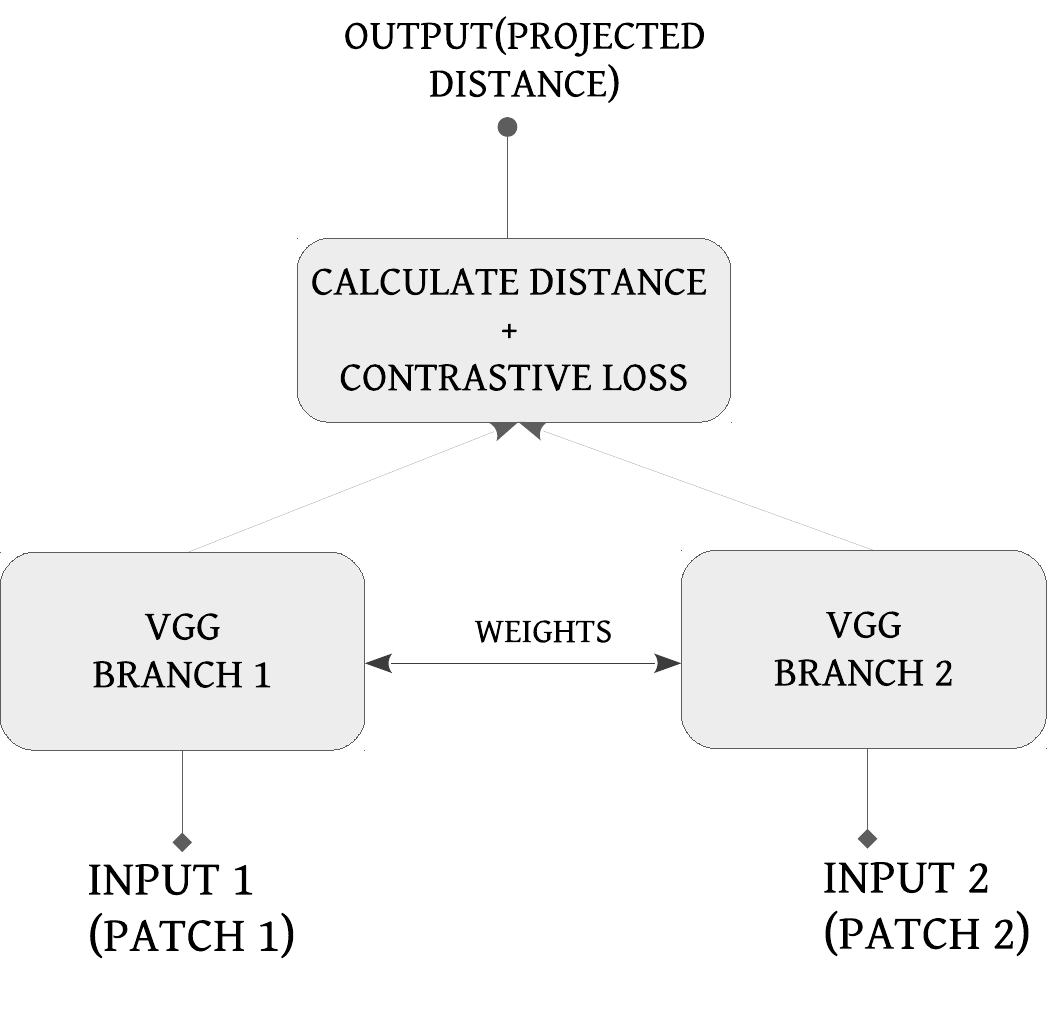}
    \caption{VGG Siamese network with contrastive loss.}
    \label{fig:network_structure_vgg_siamese}
\end{figure}

The VGG network \cite{VGG} is a CNN which was conceptualized by K. Simonyan and A. Zisserman from the University of Oxford (Visual Geometry Group). This network performed very well in ImageNet challenge 2014. The architecture/s has 
very small 3x3 Conv filters and depth varying from 16 to 19 weight layers. This network  generalizes very well with different kinds of data. VGG network has been chosen as the branches of the Siamese (Figure \ref{fig:network_structure_vgg_siamese}) network 
It's role is to extract features, similar to the DenseNet-Siamese, the final decision making and prediction is done by the Siamese network. The network is trained with Contrastive loss. The output of this network is euclidean distance between the two input sonar patches, projected into lower dimension using Contrastive loss. The hyper-parameters of this network are shown in Table \ref{table:Search_space_contrastive}.

Since contrastive loss returns projected distance, close to zero means similarity and higher values means dissimilarity. Although, in our original data and matching formulation, labels close to one represents similarity between patches. Hence the labels for train, validation and test data here are all flipped:
\begin{equation}
    Y_{\text{new}} = 1 - Y_{\text{old}}
    \label{eq:flipped_labels}
\end{equation}
Equation \ref{eq:flipped_labels} is applied to all ground truth labels, meaning that for this evaluation input label zero means similarity (match) between patches.

\begin{table}[t]
    \centering
    \caption{Best hyper-parameter values for VGG Siamese network with Contrastive loss (CL).}
    {\begin{tabular}{llll} 
            \toprule
            \multicolumn{1}{l}{\textbf{Name}} & \multicolumn{1}{l}{\textbf{Value}} & \multicolumn{1}{l}{\textbf{Name}} & \multicolumn{1}{l}{\textbf{Value}} \\ [0.5ex] 
            \midrule
            Conv filters & 16 & Kernel size & 3\\
            FC Layers & 1 &  FC output & 2048 \\ 
            Batch normalization & False & Dropout & 0.6 \\
            Batch size & 256 & Optimizer & Nadam\\
            Conv Initializer & random normal & FC Initializer & glorot normal \\
            Learning rates & 0.0002\\
            \toprule
    \end{tabular}}
    \label{table:Search_space_contrastive}
\end{table}

\section{Experimental Evaluation}

\subsection{Dataset}

We use the Marine Debris dataset, matching task, \footnote{Available at \url{https://github.com/mvaldenegro/marine-debris-fls-datasets/releases/}} to evaluate our models. This dataset contains 47K labeled sonar image patch pairs, captured using a ARIS Explorer 3000 Forward-Looking sonar, generated from the original 2627 labeled object instances. We exclusively use the \textbf{D} dataset, on which the training and testing sets were generated using different sets of objects, with the purpose of testing a truly generic image matching algorithm that is not object specific. The training set contains 39840 patch pairs, while the test set contains 7440 patch pairs.





\subsection{Comparative Analysis of AUC}

\begin{figure}[t]
    \centering
    \includegraphics[width=0.41\textwidth]{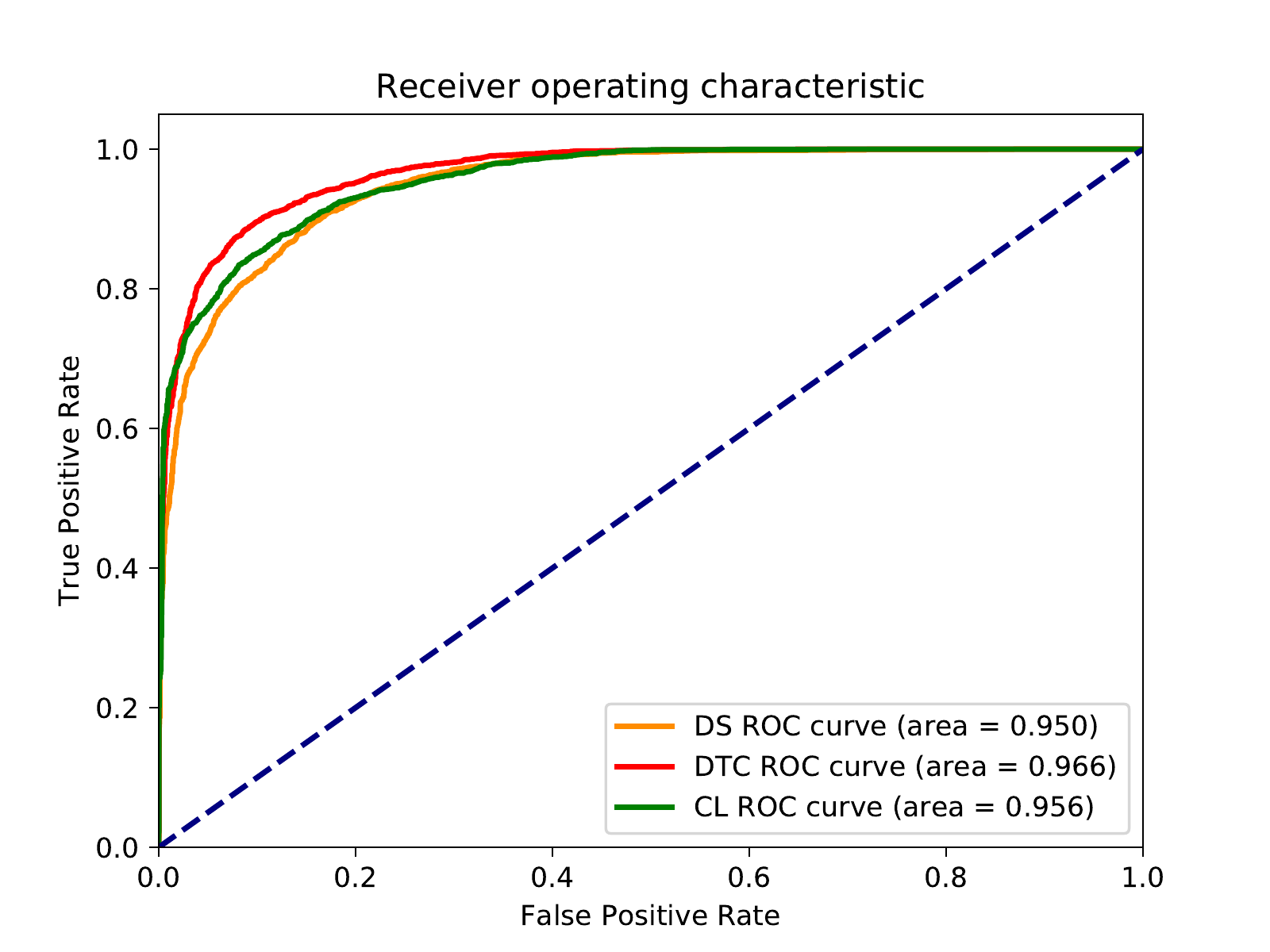}
    \caption{Comparison of ROC curves for best hyper-parameter architecture configurations and top AUC.}
    \label{fig:final_auc_compare}
\end{figure}

Our main results are presented in Table \ref{table:comparative_auc_results} and Figure \ref{fig:final_auc_compare}, where we present the AUC and the ROC curves on the test set, correspondingly.

DenseNet two-channel has highest mean AUC (10 trials) of $0.955 \pm 0.009$ with max AUC of 0.966. With total parameters of only 51,430. 
DenseNet-Siamese has highest mean AUC (10 trials) of $0.921 \pm 0.016$, Max AUC 0.95 with total parameters of 16,725,485. 
VGG-Siamese network with Contrastive loss have mean AUC (10 trials) of $0.949 \pm 0.005$ and highest AUC value in a single run as 0.956. With total number of parameters of 3,281,840. These AUC values are considerably better than Valdenegro-Toro \cite{stateoftheart}, with improvements from $0.910$ to $0.966$ (almost 5 AUC points).

It is notable that our best performing model is a two-channel network, indicating that this meta-architecture is better suited for the matching problem than a siamese one, and that there is a considerable reduction in the number of parameters, from $1.8$M to $51$K, which hints at increased generalization.

\begin{figure*}[!tb]
    \centering
    \includegraphics[width=0.95\textwidth]{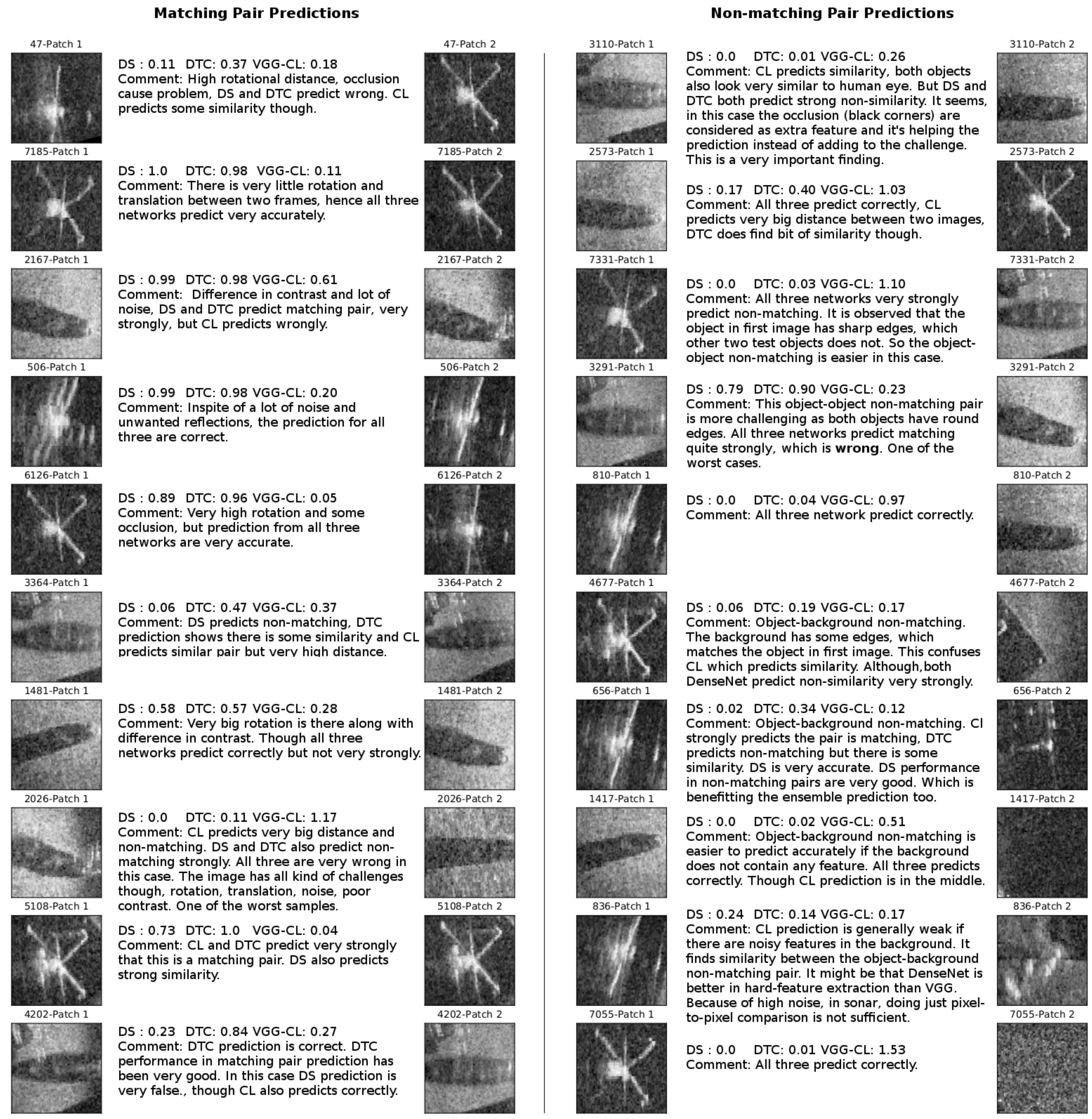}
    \caption{Comparison of predictions across multiple models, DenseNet Siamese (DS), DenseNet Two-Channel (DTC), and VGG Siamese Contrastive Loss (CL). Note that Siamese VGG produce distances which are not in the range $[0, 1]$, while the other architectures give scores in the $[0, 1]$ range.}
    \label{fig:prediction_multimodel_comparison}
\end{figure*}

A comparison of predictions between all our three architectures is provided in Figure \ref{fig:prediction_multimodel_comparison}.

\begin{table}[t]
    \centering
    \caption{Comparative analysis on the AUC and total number of parameters in the best performing networks.}
    {\begin{tabular}{llll} 
            \toprule
            \textbf{Network} 				& \textbf{AUC} 				&  \textbf{Best AUC} 	& \textbf{\# of Params}\\
            \midrule
            Two-Channel DenseNet 	& $0.955 \pm 0.009$ & 0.966 	& 51K\\
            Siamese DenseNet 		& $0.921 \pm 0.016$ & 0.95 		& 16.7M \\
            Siamese VGG			& $0.949 \pm 0.005$ & 0.956 	& 3.3M\\
            \midrule
            Two-Channel CNN \cite{stateoftheart}		& 0.910				& 0.910		& 1.8M \\
            Siamese CNN \cite{stateoftheart}		& 0.855				& 0.855		& 1.8M \\
            \toprule
    \end{tabular}}
    \label{table:comparative_auc_results}
\end{table}

\subsection{Monte Carlo Dropout Analysis}
Normally Dropout is only applied in the training phase, where it provides regularization to avoid overfitting. In test time all the connections/nodes remain present and dropout is not applied, though the weights are adjusted according to the dropout ratio during training. So every time a prediction on test data is obtained, they are deterministic.
For Monte Carlo dropout the dropout is also applied in the inference/test time, which introduces randomness, as the connections are dropped randomly according to the dropout probability. This prediction process is stochastic i.e the model could predict 
different predictions for same test data. The main goal of Monte Carlo Dropout \cite{Gal2015DropoutB} is to generate samples of the predictive posterior distribution of an equivalent Bayesian Neural Wetwork, which quantifies epistemic uncertainty.

We would like to evaluate uncertainty for our best performing model, the DenseNet two-channels (AUC 0.966). This model is trained with Dropout with $p = 0.2$. For this evaluation the MC-Dropout during inference time is enabled explicitly. 20 forward passes for each of the test images are made and the mean score and standard deviation is computed. The standard deviation is a measure of uncertainty, with increasing value indicating more uncertainty.

\begin{figure}[t]
    \centering
    \begin{subfigure}[b]{0.32\columnwidth}
        \includegraphics[width=\textwidth]{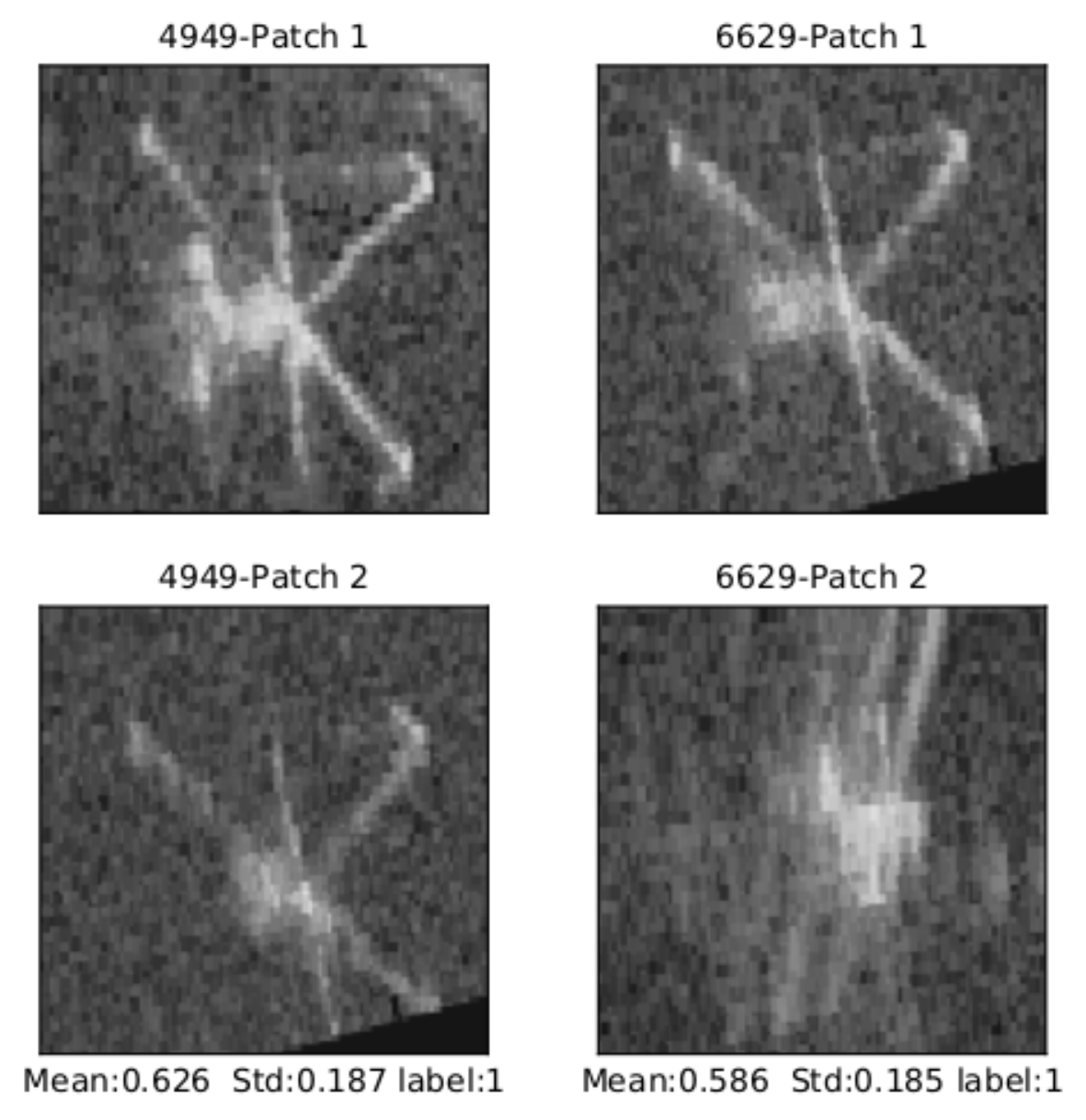}
        \label{fig:prediction_images_highest_stds_1}
    \end{subfigure}
    \begin{subfigure}[b]{0.32\columnwidth}
        \includegraphics[width=\textwidth]{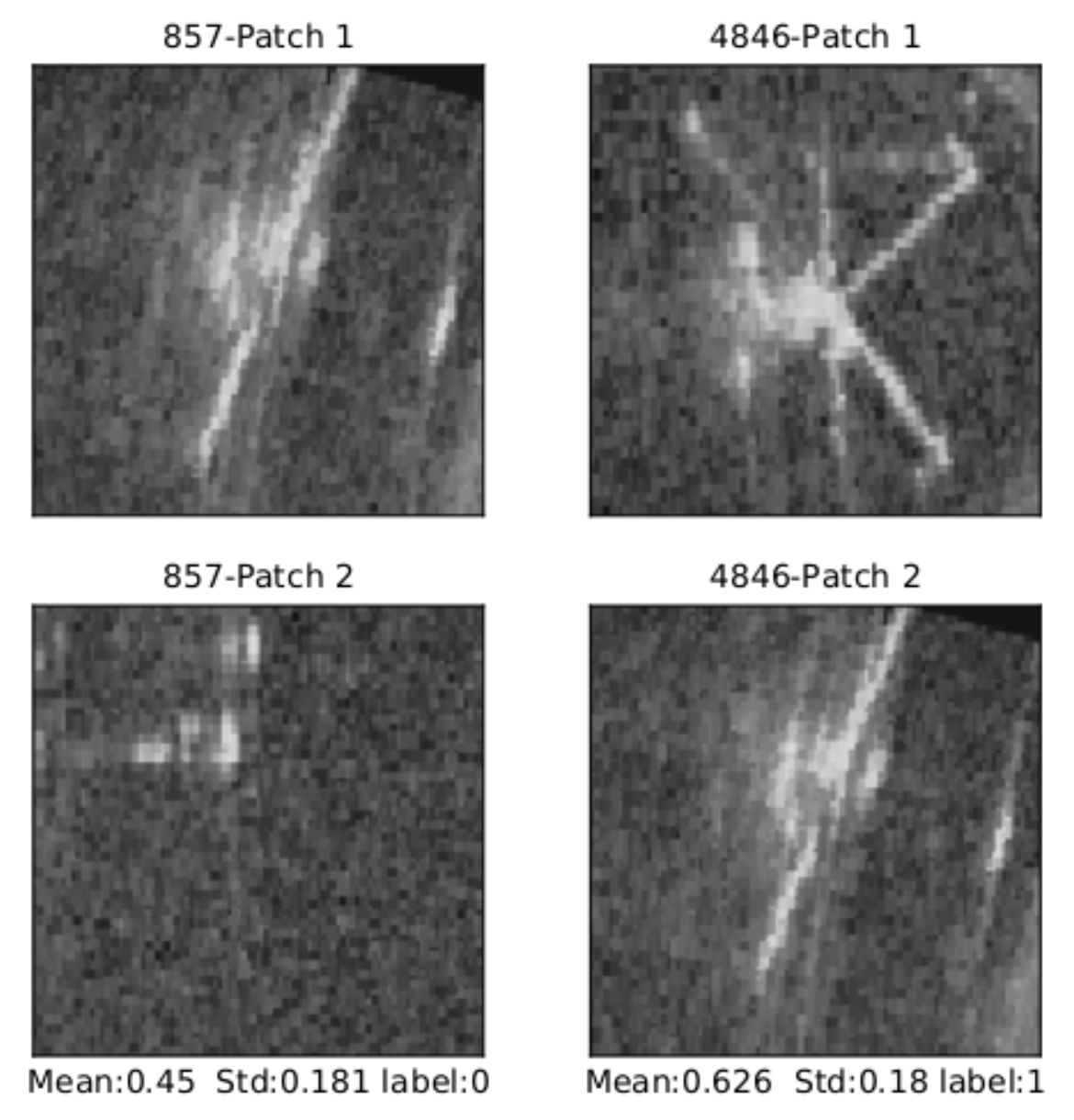}
        \label{fig:prediction_images_highest_stds_2}
    \end{subfigure}   
    \begin{subfigure}[b]{0.32\columnwidth}
        \includegraphics[width=\textwidth]{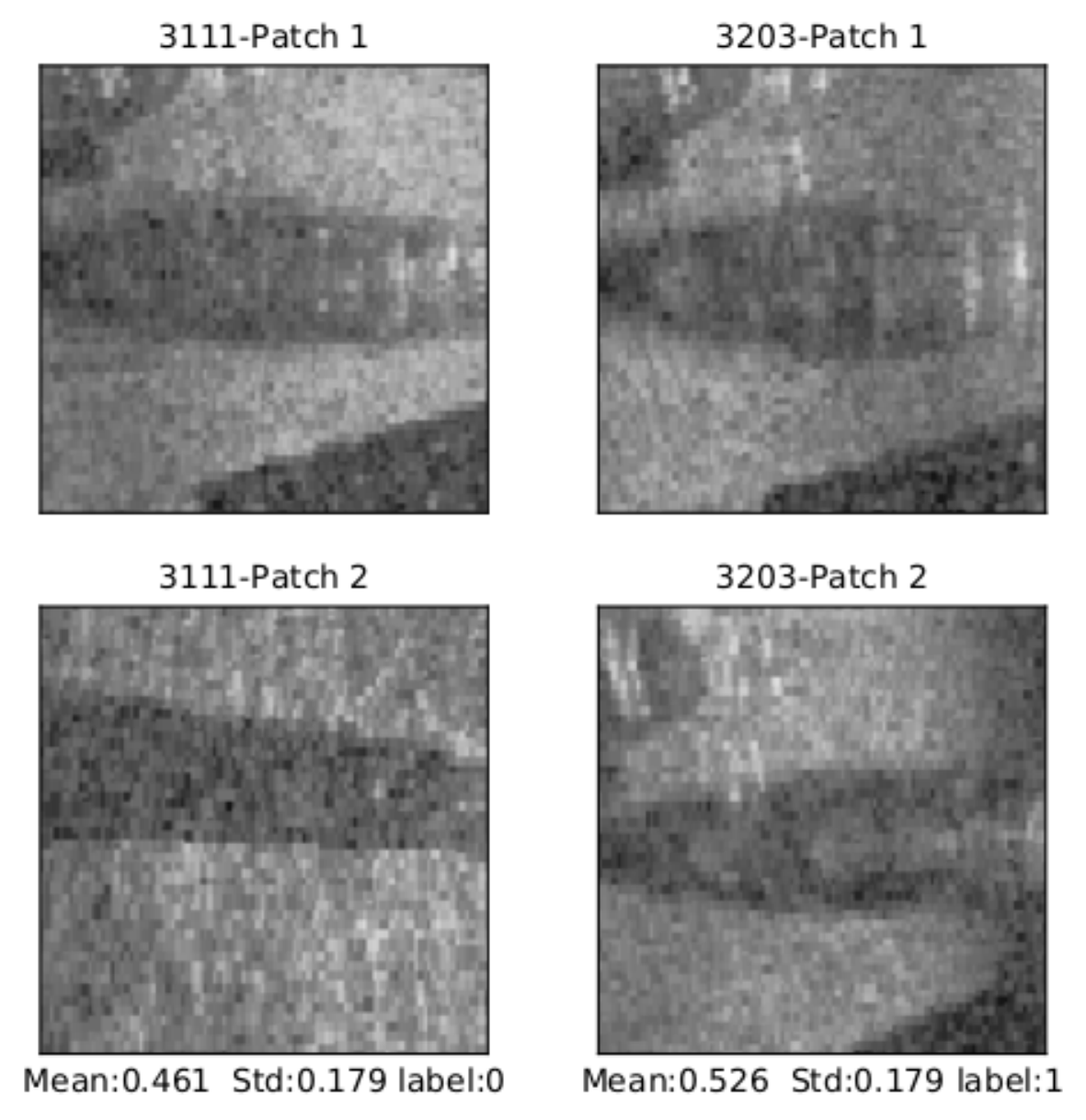}
        \label{fig:prediction_images_highest_stds_3}
    \end{subfigure} 
    \caption{MC-Dropout predictions of DTC with highest standard deviation over 20 forward passes. Ground truth label 1 indicated matching. It is clear that the low signal-to-noise for sonar is affecting the predictions, and unwanted reflections and occlusions are also challenging.}
    \label{fig:mc_prediction_highest_std}
\end{figure}

\begin{figure}[t]
    \centering
    \begin{subfigure}[b]{0.32\columnwidth}
        \includegraphics[width=\textwidth]{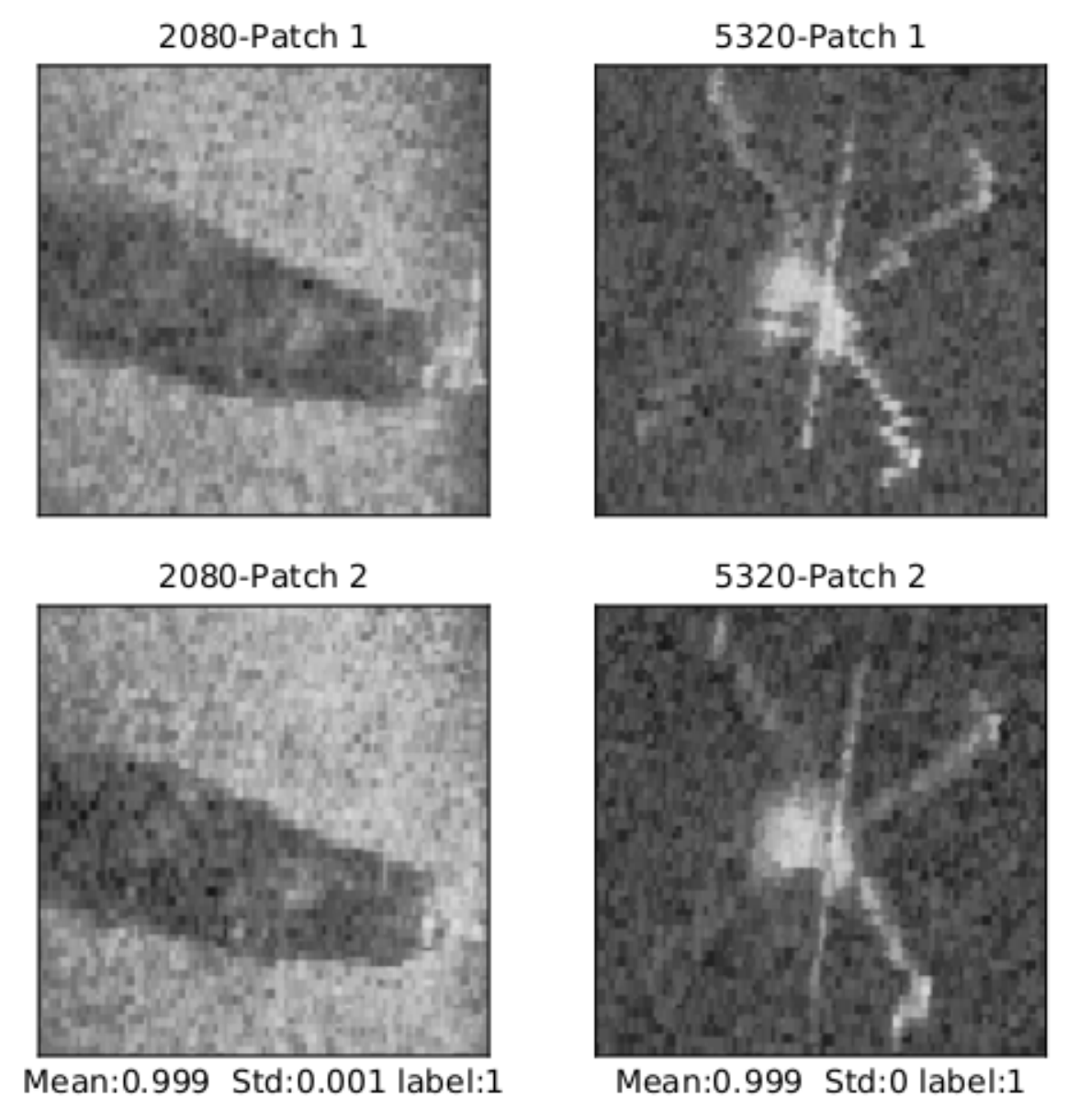}
        \label{fig:prediction_images_lowest_stds_1}
    \end{subfigure}
    \begin{subfigure}[b]{0.32\columnwidth}
        \includegraphics[width=\textwidth]{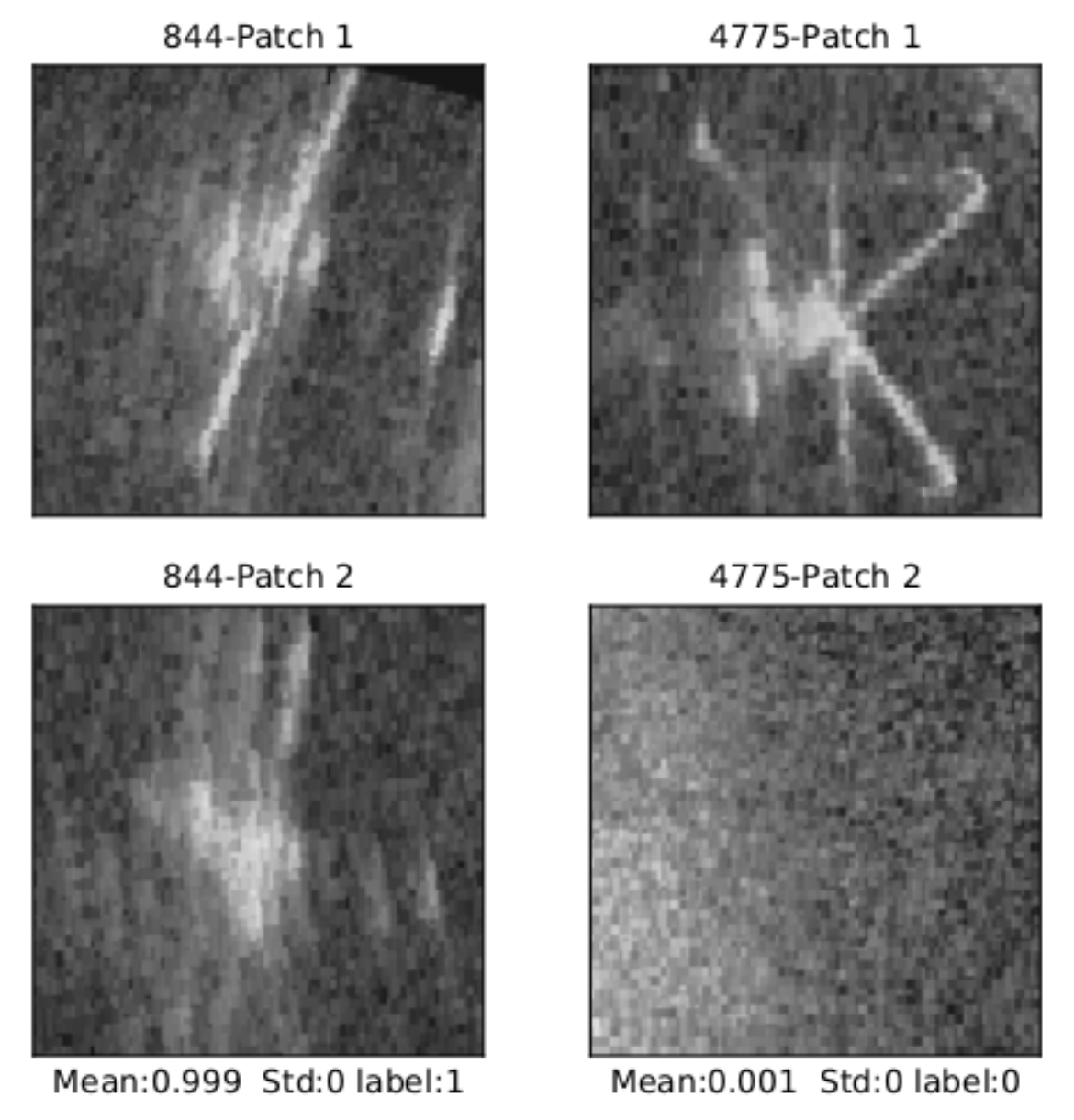}
        \label{fig:prediction_images_lowest_stds_2}
    \end{subfigure}  
    \begin{subfigure}[b]{0.32\columnwidth}
        \includegraphics[width=\textwidth]{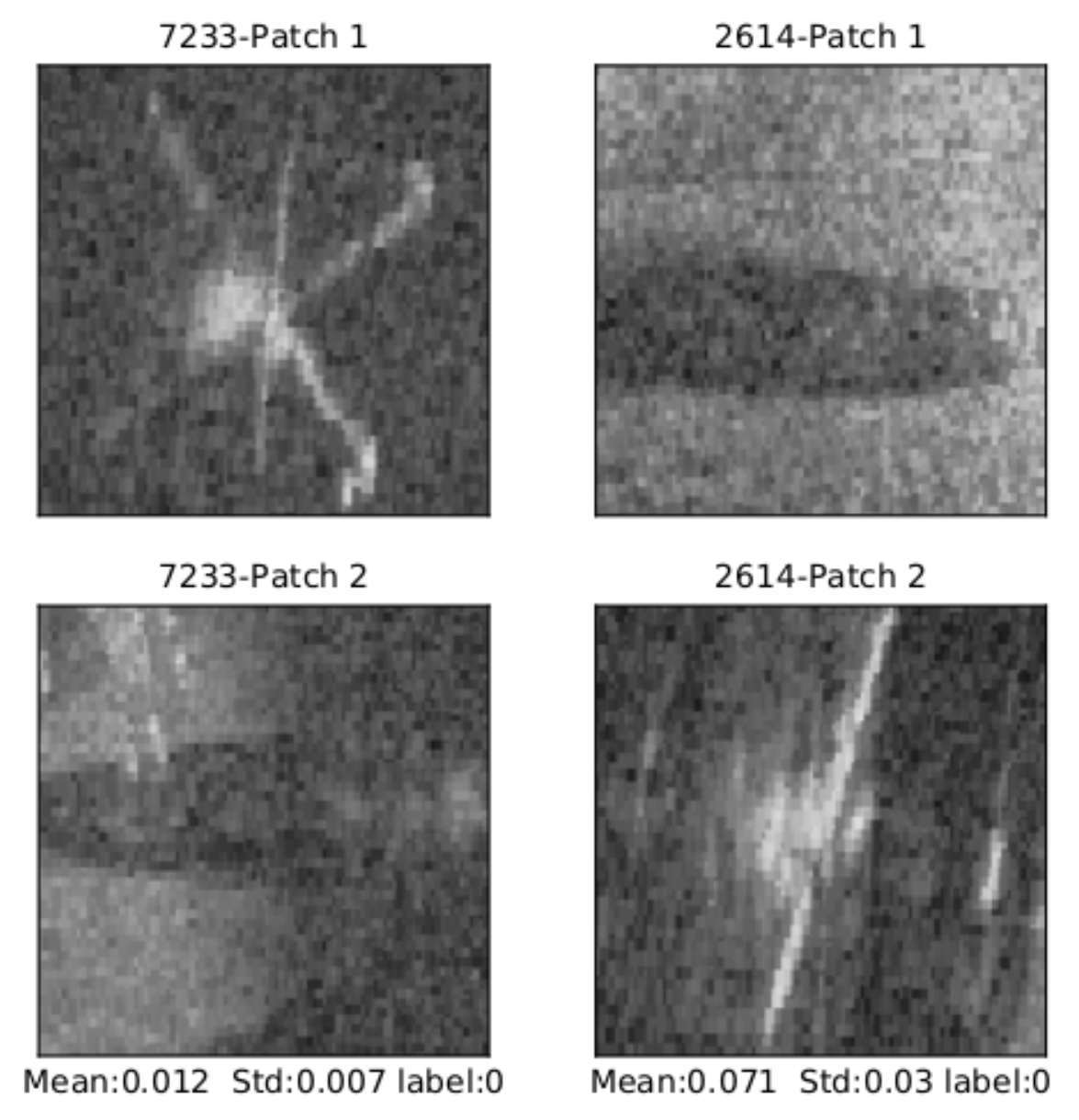}
        \label{fig:prediction_images_lowest_stds_3}
    \end{subfigure} 
    \caption{MC-Dropout predictions of DTC with lowest standard deviation over 20 forward passes. These results show that the network learned some of the similarity functions with great confidence. For object-object non-matching pairs usual std values are much higher than other categories.}
    \label{fig:mc_prediction_lowest_std}
\end{figure}

Figures \ref{fig:mc_prediction_highest_std} and \ref{fig:mc_prediction_lowest_std} present these results in terms of the most uncertain patch pairs in Figure \ref{fig:mc_prediction_highest_std}, and the most certain (least uncertain) images in Figure \ref{fig:mc_prediction_lowest_std}. These results give insights on what the model thinks are its most difficult samples (high uncertainty), and in particular, the most uncertain examples (highest standard deviation) are the ones close to being out of distribution, where the patches are positioned near the border of the FLS polar field of view, which probably confuses the model.

The lowest uncertainty results in Figure \ref{fig:mc_prediction_lowest_std} indicate the easiest patch pairs to discriminate, either the same object in relatively similar poses, or radically different objects or background in each pair. In both cases the model is quite confident of these predictions.

\begin{figure*}[!tb]
    \centering
    \includegraphics[width=0.785\textwidth]{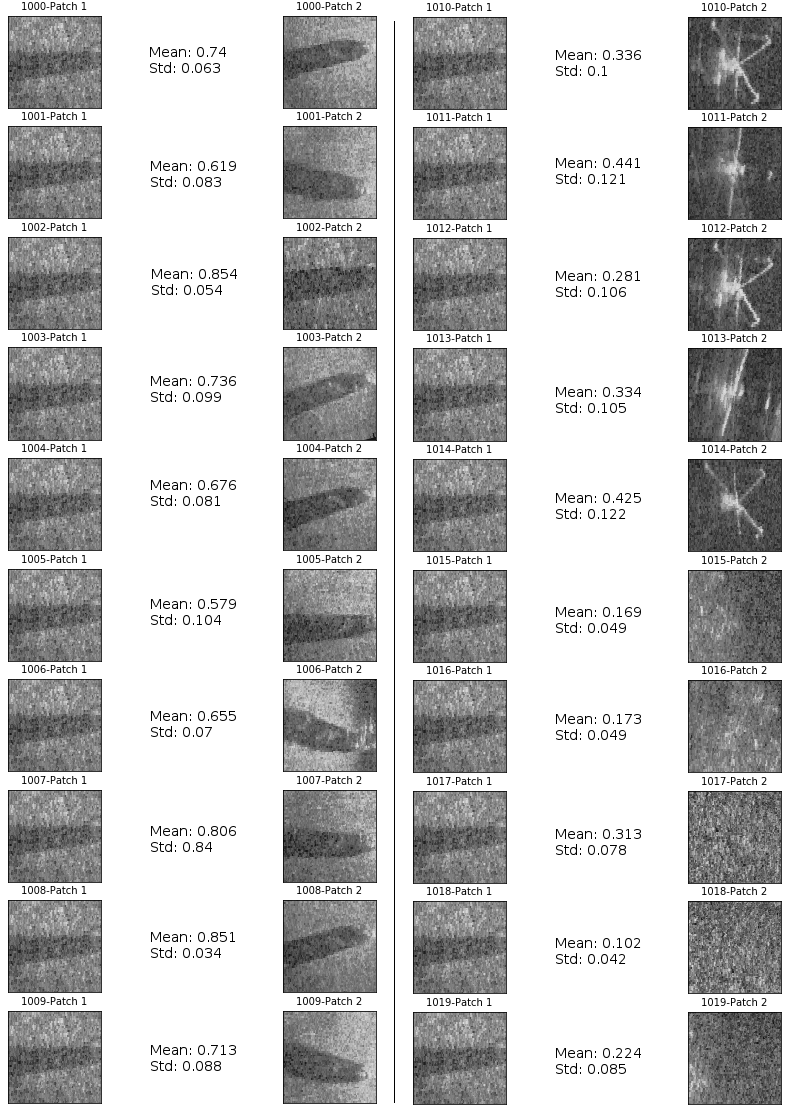}
    \caption{MC-Dropout predictions of DTC with highest standard deviation over 20 forward passes, presenting 20 sonar image patch pairs from the test dataset with index 1000 to 1019, and corresponding mean prediction and standard deviation.}
    \label{fig:prediction_images_MC}
\end{figure*}

Figure \ref{fig:prediction_images_MC} shows a large selection of patch pairs and their uncertainty estimates, showing that the model is not always confident, particularly for predictions with scores in between zero and one, even for pairs that a human would consider to be easy to match or reject.

\subsection{Ensemble}
The performance of the DenseNet-Siamese(DS) is good for non-matching pair predictions. DenseNet two-channel(DTC) is overall very good, but most uncertain in object-object non matching pairs.

This observation led to the hypothesis that making an ensemble of these two classifiers might improve overall predictive capability. For this experiment a few of the previously trained models of DTC and DS are loaded, and their predictions on the test data are averaged,i.e. same weights for DS and DTC both. These evaluation results are displayed in Table \ref{table:Ensemble}. The ROC AUC calculated on the average prediction is found to be higher than the individual scores each time.

\begin{table}[t]
    \centering
    \caption{After combining DS and DTC models, with AUC presented in first two columns, the Ensemble is encoded and its prediction accuracy (AUC) gets much improved, presented in the third column.}
    \resizebox{0.45\textwidth}{!}
    {\begin{tabular}{lll} 
            \toprule
            \textbf{DS model AUC} &  \textbf{DTC model AUC} &  \textbf{Ensemble AUC}\\ [0.5ex]
            \midrule
            0.95 & 0.959 & 0.97\\
            0.952 &  0.959 &  0.97\\
            0.952 &  0.963 &  0.973\\
            0.952 &  0.966 &  0.971\\
            0.952 &  0.972 &  0.978\\
            \toprule
    \end{tabular}}
    \label{table:Ensemble}
\end{table}

Ensemble accuracies (AUC) are consistently better than each model individually. If the underlying models, which encode the ensemble, has low AUC, the ensemble AUC is found to be much-improved. For example the first result presented in 
Table \ref{table:Ensemble} where the ensemble accuracy is much higher (0.97 AUC) than the underlying model predictions (0.95 and 0.959 AUCs). By encoding an ensemble of the DenseNet-Siamese model with AUC 0.952 and the DenseNet two-channel
model with 0.972 AUC, the resulting ensemble AUC is found to be 0.978, which is the highest AUC on test data obtained in any other experiment during the scope of this work. This indicates that both DS and DTC models are complementary and could be used together if higher AUC is required in an application.

\section{Conclusions and Future Work}
In this work we present new neural network architectures for matching of sonar image patches, including extensive hyper-parameter tuning, and explore their performance in terms of area under the ROC curve, uncertainty as modeled by MC-Dropout, and performance as multiple models are ensembles. The results in this work are proven to be improvements over the state of the art on the same dataset. Using DenseNet two-Channel network, average prediction accuracy obtained is 0.955 area under ROC curve (AUC). VGG-Siamese (with Contrastive loss function) and DenseNet-Siamese perform the prediction with an average AUC of 0.949 and 0.921 respectively. All these results are an improvement over the result of 0.910 AUC from Valdenegro-Toro \cite{stateoftheart}. Furthermore, by encoding an ensemble of DenseNet two-channel and DenseNet-Siamese models with respective highest AUC scores, prediction accuracy for the Ensemble obtained is 0.978 AUC, which is overall highest accuracy obtained in the Marine Debris Dataset for the matching task.

We expect that our results motivate other researchers to build applications on top of our matching networks.

\bibliographystyle{IEEEtran}
\bibliography{ecai}

\end{document}